\documentclass[letterpaper, 10 pt, conference]{ieeeconf}

\IEEEoverridecommandlockouts
\usepackage{algorithmic}
\usepackage{algorithm}
\usepackage{amssymb,amsmath}
\usepackage{roboticsMathNotations}
\usepackage{xcolor}
\usepackage{url}
\usepackage{multicol}
\usepackage{multirow}
\usepackage[utf8x]{inputenc}
\usepackage[english]{babel}
\usepackage{algorithmic}
\usepackage{algorithm}
\usepackage{subfigure}
\usepackage{siunitx}
\usepackage{caption}
\usepackage{tabularx}
\usepackage{booktabs}

\newcolumntype{Y}{>{\raggedleft\arraybackslash}X}
\newcolumntype{b}{Y}
\newcolumntype{z}{>{\hsize=.55\hsize}Y}
\newcolumntype{s}{>{\hsize=.15\hsize}Y}

\makeatletter
\newcommand{\todolist}[1]{\begin{itemize}\item \textcolor{red}{#1}\checknextarg}
	\newcommand{\checknextarg}{\@ifnextchar\bgroup{\gobblenextarg}{\end{itemize}}}
\newcommand{\gobblenextarg}[1]{ \item \textcolor{red}{#1}\@ifnextchar\bgroup{\gobblenextarg}{\end{itemize}}}

\newcommand{\frameRobot}{R}
\newcommand{\frameOdom}{O}
\newcommand{\frameCamera}{C}
\newcommand{\frameMarker}{M}
\newcommand{\framePanel}{P}
\newcommand{\robotPoseSimulation}{\ensuremath{\trfFr{\frameRobot}{\frameOdom}{_{S}}}}
\newcommand{\robotPoseMarker}{\ensuremath{\trfFr{\frameRobot}{\frameOdom}{_{M}}}}
\newcommand{\robotPoseVO}{\ensuremath{\trfFr{\frameRobot}{\frameOdom}{_{VO}}}}
\newcommand{\robotPoseFilter}{\ensuremath{\trfFr{\frameRobot}{\frameOdom}{_{F}}}}

\newcommand{\robotPoseError}[2]{\ensuremath{d\left(\trfFr{\frameRobot}{\frameOdom}{_{#1}},\trfFr{\frameRobot}{\frameOdom}{_{#2}}\right)}}
\newcommand{\robotPositionError}[2]{\ensuremath{d\left(\posFrMean{\frameRobot}{\frameOdom}{_{#1}},\posFrMean{\frameRobot}{\frameOdom}{_{#2}}\right)}}
\newcommand{\robotOrientationError}[2]{\ensuremath{d\left(\quatFrMean{\frameRobot}{\frameOdom}{_{#1}},\quatFrMean{\frameRobot}{\frameOdom}{_{#2}}\right)}}

\newcommand{\panelPositionError}[2]{\ensuremath{d\left(\posFrMean{\framePanel}{\frameOdom}{_{#1}},\posFrMean{\framePanel}{\frameOdom}{_{#2}}\right)}}
\newcommand{\panelOrientationError}[2]{\ensuremath{d\left(\quatFrMean{\framePanel}{\frameOdom}{_{#1}},\quatFrMean{\framePanel}{\frameOdom}{_{#2}}\right)}}

\usepackage{pgf}
\newcommand\inputpgf[2]{{
		\let\pgfimageWithoutPath\pgfimage
		\renewcommand{\pgfimage}[2][]{\pgfimageWithoutPath[##1]{#1/##2}}
		\input{#1/#2}
}}
\usepackage{adjustbox}

\title{\bf Adaptive Navigation Scheme for Optimal Deep-Sea Localization Using Multimodal Perception Cues}
\author{Arturo Gomez Chavez$^{1}$, Qingwen Xu$^{2}$, Christian A. Mueller$^{1}$, S\"oren Schwertfeger$^{2}$ and Andreas Birk$^{1}$ % <-this % stops a space
	\thanks{$^{1}$Authors are with the Robotics Group, Computer Science \& Electrical Engineering, Jacobs University Bremen, Germany.
		{\tt\small \{a.gomezchavez, chr.mueller, a.birk\}@jacobs-university.de}}%
	\thanks{$^{2}$Authors are with the School of Information Science Technology of ShanghaiTech University
		{\tt\small <xuqw, soerensch>@shanghaitech.edu.cn}}%
	\thanks{* This research received funding from the European Union's Horizon 2020 Framework Programme, project ref. 635491: ``Effective dexterous ROV operations in presence of communication latencies (DexROV)''.}
}

\begin{document}
\maketitle
\bstctlcite{IEEEexample:BSTcontrol}

\begin{abstract}
	Underwater robot interventions require a high level of safety and reliability. 
	A major challenge to address is a robust and accurate acquisition of localization estimates, as it is a prerequisite to enable more complex tasks, e.g. floating manipulation and mapping.
	State-of-the-art navigation in commercial operations, such as oil\&gas production (OGP), rely on costly instrumentation. These can be partially replaced or assisted by visual navigation methods, especially in deep-sea scenarios where equipment deployment has high costs and risks.
	Our work presents a multimodal approach that adapts state-of-the-art methods from on-land robotics, i.e., dense point cloud generation in combination with plane representation and registration, to boost underwater localization performance.
	A two-stage navigation scheme is proposed that initially generates a coarse probabilistic map of the workspace, which is used to filter noise from computed point clouds and planes in the second stage. 
	Furthermore, an adaptive decision-making approach is introduced that determines which perception cues to incorporate into the localization filter to optimize accuracy and computation performance. 
	Our approach is investigated first in simulation and then validated with data from field trials in OGP monitoring and maintenance scenarios.   
\end{abstract}

%=== INTRODUCTION ===%
\section{Introduction}
\label{introduction}

The marine environment is challenging for automation technologies.
Yet, oceans are one of the main forces driving commerce, employment, economic revenue and natural resources exploitation,
which in turn triggers profound interest in the development of new technologies to facilitate intervention tasks, e.g., in oil\&gas production (OGP).
Remote Operated Vehicles (ROVs) are the current work-horse used for these tasks, which include inspection of ships, submerged structures and valves manipulation. 

In particular, manipulation tasks are extremely challenging without stable and accurate robot localization.
In general, a global positioning based navigation is desirable to correct measurements from inertial navigation systems (INS) in a \emph{tightly-coupled} approach~\cite{Tal2017_uwnavigation}.
However, such data has to be transmitted acoustically through ultra-short/long baseline (USBL/LBL) beacons that have low bandwidth~\cite{Stutters2008_uwnavigation}, signal delays and deployment constraints.
Additional sensors, i.e., Doppler velocity logs (DVLs) and digital compasses, can improve the localization accuracy but still not at the required standards to perform \emph{floating-base} manipulation.
%But especially high-end devices are very costly, and it still challenging to achieve the required accuracy to perform \emph{floating-base} manipulation.

\begin{figure}
	\small
	\centering
	\subfigure[]{\label{fig:rov_and_panel}
		\includegraphics[width=0.95\linewidth]{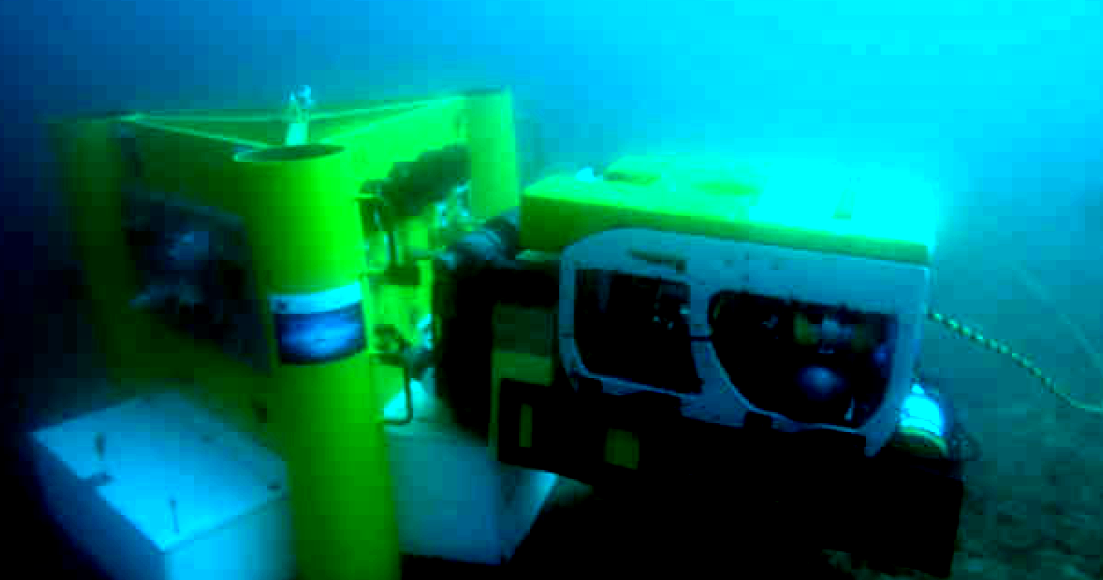}}
	\subfigure[]{\label{fig:rov_stereo_camera}
		\includegraphics[width=0.395\linewidth]{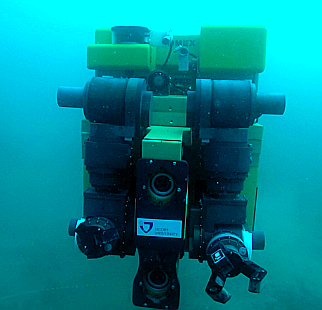}}
	\subfigure[]{\label{fig:panel_analog_camera}
		\includegraphics[width=0.555\linewidth]{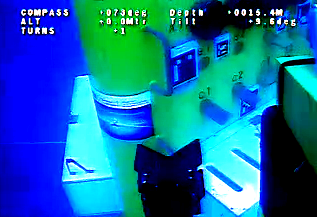}}
	\caption{\subref{fig:rov_and_panel} ROV performing oil\&gas valve manipulation tasks. \subref{fig:rov_stereo_camera} ROV stereo camera system and manipulation arms. \subref{fig:panel_analog_camera} ROV first person view while approaching oil\&gas panel.}
	\label{fig:dexrov_in_action}
\end{figure}

We present a navigation scheme that uses visual odometry (VO) methods based on stereo camera imagery and an initial probabilistic map of the working space to boost localization accuracy in challenging conditions.
The application scenario is the monitoring and dexterous manipulation of a OGP panel (Fig.~\ref{fig:dexrov_in_action}) within the EU-project DexROV~\cite{Mueller2018_DexROVSIL,Birk2018}%,Gancet2016}.
%However, the nature of underwater scenarios where the light behavior produces low contrast, blurred and color attenuated images highly impacts the performance of VO approaches that rely on image features.

To address the challenges of underwater vision, we combine plane registration and feature tracking methods.
3D planes are extracted from dense point cloud (DPC) generators, which produce complete disparity maps at the cost of depth accuracy; density being the key factor to find reliable 3D planes.
This is particularly useful in structured or man-made environments, which predominantly contain planar surfaces, like the installations used in OGP. 
Furthermore, a decision-making strategy based on image quality is introduced: it allows to select the visual odometry method to be used in order to obtain reliable measurements and improve computation times.
%Last but not least, a contribution of this paper is the validation of the work in realistic field conditions with data from sea trials off-shore the coast of Marseille, France.
In summary our contributions are:
\begin{itemize}
	\item Development of different visual odometry (VO) modalities based on: knowledge-enabled landmarks, 3D plane registration and feature tracking.
	\item Integration of the multimodal VO into an underwater localization filter that adapts its inputs based on image quality assessments. 
	\item A two-step navigation scheme for structured environments. Initial suboptimal localization measurements are used to compute a coarse probabilistic 3D map of the workspace. In turn, this map is used to filter noise and optimize the integration of further measurements.
	%first, \emph{loose} localization criteria is used to compute a coarse probabilistic 3D map of the workspace. Second, which in turn is later used to optimize localization measurements.
	\item Validation of the presented scheme in realistic field conditions with data from sea trails off-shore the coast of Marseille, France.
\end{itemize}

%Given a related work summary in Sec.~\ref{sec:related work}, in Sec.~\ref{sec:methodology intro} our methodology is described including \emph{knowledge-enabled localization}, our \emph{visual odometry} approach derived from \emph{dense depth mapping} and \emph{plane extraction} methods; plus the proposed \emph{image quality assessment} for selective navigation input modality. 
%An experimental evaluation is conduced in Sec.~\ref{sec:experiments}, followed by concluding statements in Sec.~\ref{sec:conclusion}.

%%=== RELATED WORK ===%
\section{Related Work}
\label{sec:related work}

A great number of theoretical approaches on localization filters for marine robotics have been proposed in the literature.
In recent years, this also includes increasing efforts to address practical issues such as multi-rate sampling, sensor glitches and dynamic environmental conditions.
In \cite{Paull2014_uwnavigation}, a review of the state-of-the-art in underwater localization is presented and classified into three main classes: inertial/dead reckoning, acoustic and geophysical.
 
The surveyed methods show a clear shift from technologies like USBL/LBL positioning systems~\cite{Morgado2013_usbl} towards two research areas. First, dynamic multiagent systems which include a surface vehicle that complements the underwater vehicles position with GPS data~\cite{Campos2016_multivehicle}; and secondly, the integration of visual navigation techniques, i.e., visual odometry~\cite{Sukvichai2016_auvvo} and SLAM~\cite{Fallon2013_sonarslam}, into marine systems.  
%
%The system and approach presented in this article fall under the \emph{inertial} and \emph{geophysical} sensor category since we use external environmental information as a reference for localization. 
%One reason is that deep-sea technology makes the use of multi-vehicle systems highly expensive at the moment.
We also integrate inertial data from DVL and IMU with vision-based techniques using standard 2D features and in addition 3D plane registration. 
The work in~\cite{Proenca2017_rgbdodometry} shows that the combination of standard visual features with geometric visual primitives increases odometry robustness in low texture regions, highly frequent in underwater scenes.

Three methods are commonly used for plane primitive extraction: RANSAC, Hough transform and Region Growing (RG). 
State of the art methods~\cite{poppinga2008fast,Feng2014_PEAC,Proenca2018_CAPE} often use RG because it exploits the connectivity information of the 3D points and, thus, have more consistent results in the presence of noise.
These are better suited for our application since the input point cloud for the plane extraction algorithm is not directly generated from an RGB-D camera but from a stereo image processing pipeline. 
We compare some of these stereo pipelines to investigate their impact on the overall localization accuracy (see Sec.~\ref{exp: dense maps}).

Finally, to test the complete framework, we used the \emph{continuous system integration and validation} (CSI) architecture proposed in our previous work~\cite{Fromm2017}. 
%Finally, to test the complete framework, we use a \emph{continuous system integration and validation} (CSI) architecture~\cite{Fromm2017}. 
With this architecture, parts of the developmental stack can be synchronized with real-world data from field trials to close the discrepancy between simulation and the real world;
%in ~\cite{Mueller2018_DexROVSIL} we denominate this process as \emph{simulation in the loop} (SIL) methodology.
this process is inspired by the \emph{simulation in the loop} (SIL) methodology~\cite{Mueller2018_DexROVSIL}.

Based on this, we first compute the accuracy of our approach in an optimized simulation environment reflecting similar light conditions as observed in underwater trials. 
Then, its effectiveness is validated on field trial data featuring real-world environmental conditions.

%
%%=== METHODOLOGY ===%
\section{Methodology}
\label{sec:methodology intro}
\begin{figure*}[tb]
	\small
	\centering
	\includegraphics[width=0.94\linewidth]{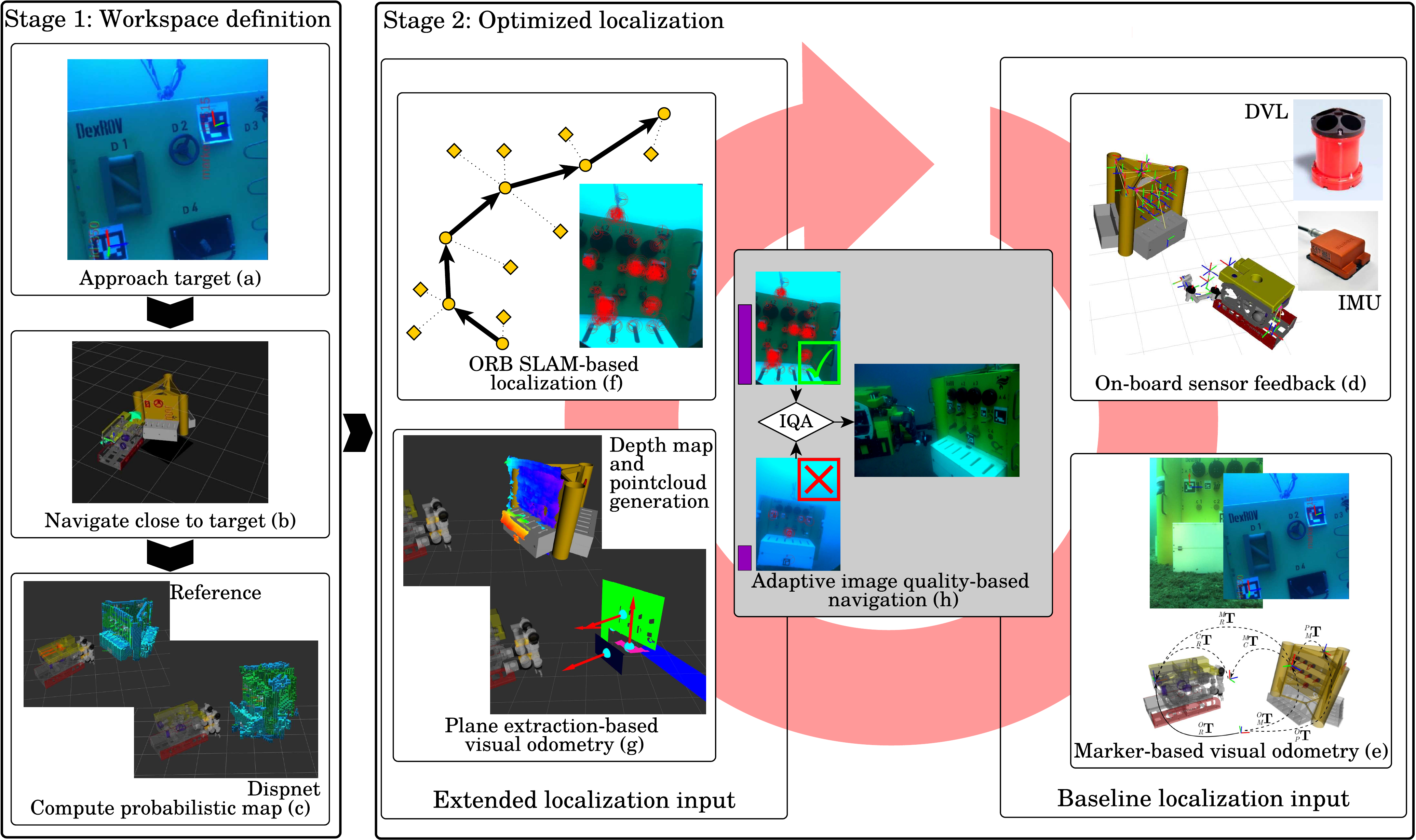}
	\caption{Proposed two-stage navigation scheme. \textbf{First stage} -- \textit{Workspace definition}: (a) recognize the target and compute its pose based on visual markers, (b) navigate close to the target based on navigational sensors and visual markers, (c) generate probabilistic map with stereo imagery and Dispnet; RGB-D camera based probabilistic map displayed for reference. \textbf{Second stage} -- \textit{Optimized localization}: (f)-(g) multimodal localization inputs which are incorporated to a final Kalman filter-based localization estimate. An image quality assessment ({IQA}) is introduced (h) to validate reliability of the extended localization inputs to boost the accuracy of the estimates given by the baseline inputs (see Sec.~\ref{sec: adaptive_navigation_scheme}).}
	\label{fig:overview_states_localization}
\end{figure*}
%\begin{figure}[!b]
%	\small
%	\centering
%	\subfigure[Recognize the target and compute its pose based on visual markers, Fig.~\ref{fig:rov_panel_tf}.]{\label{fig:first_stage_marker}
%		\includegraphics[height=0.4\linewidth]{fig/localization/first_stage/marker_detection.png}}
%	\subfigure[Navigate close to the target based on navigational sensors and visual markers.]{\label{fig:first_stage_navigate}
%		\includegraphics[height=0.4\linewidth]{fig/localization/first_stage/circle_around.png}}
%	\subfigure[Generated probabilistic map; RGB-D camera used for reference.]{\label{fig:first_stage_simulated_octomap}
%		\includegraphics[height=0.38\linewidth]{fig/localization/first_stage/rov_simulated_octomap.png}}
%	\subfigure[Generated probabilistic map with stereo imagery and Dispnet.]{\label{fig:first_stage_real_octomap}
%		\includegraphics[height=0.38\linewidth]{fig/localization/first_stage/rov_real_octomap.png}}
%	\caption{First stage of proposed methodology: Workspace definition with \emph{rough localization.}}
%	\label{fig:first_stage_localization}
%\end{figure}
Figure~\ref{fig:overview_states_localization} shows the proposed two-stage navigation scheme:
\newline

%Based on this, the pipeline follows the next steps in two stages:
%\begin{itemize}
	\textbf{First stage}-\emph{Workspace definition with loose localization}
%\end{itemize}
\begin{description}
	\item[1.1.] Approach the target (oil\&gas panel) until its global 3D pose its confidently determined based on a priori knowledge; see Sec.~\ref{sec: knowledge-enable localization}, Fig.~\ref{fig:overview_states_localization}(a). % and~\ref{fig:rov_panel_tf}.
	\item[1.2.] Navigate close to the target using odometry from navigation sensors and visual landmarks (\emph{baseline localization}); see Sec.~\ref{sec:marker localization} and Fig.~\ref{fig:overview_states_localization}(a)(b).
	\item[1.3.] Compute a probabilistic map from stereo input of the target object/area while navigating based on the odometry uncertainty; see examples in Fig.~\ref{fig:overview_states_localization}(c). 
\end{description}

\begin{figure}[tb]
	\small
	\centering
	\includegraphics[width=0.98\linewidth]{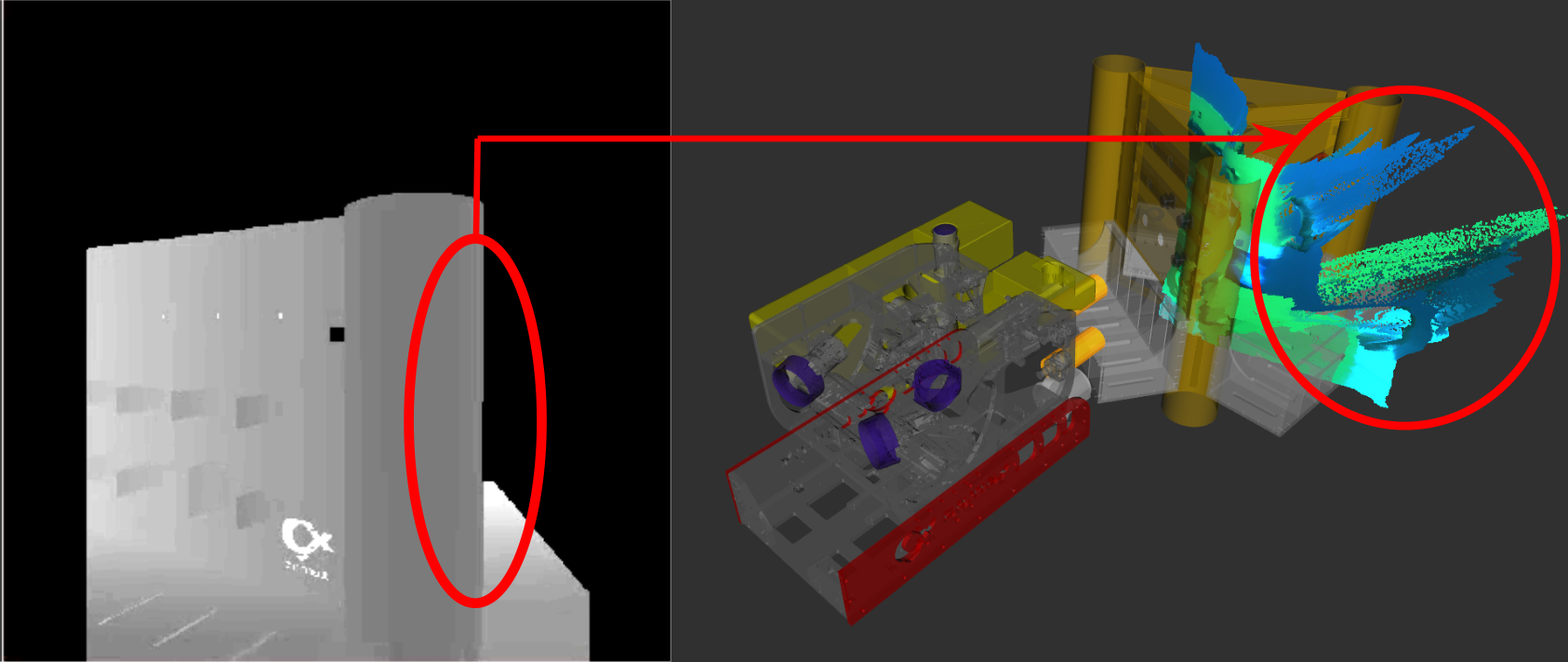}
	\caption{Comet-like artifacts (right) produced in 3D pointclouds by noisy depth maps (left). These are further filtered by the probabilistic map generated in stage 1 (Fig.~\ref{fig:overview_states_localization}(c)) to later extract planes and obtain visual odometry (Fig.~\ref{fig:overview_states_localization}(g)). }
	\label{fig:second_stage_pc_comet}
\end{figure}

%\begin{itemize}
	\textbf{Second stage}-\emph{Optimized localization}
%\end{itemize}
\begin{description}
	\item[2.1] Evaluate the reliability of the visual input, i.e., stereo image quality (Sec.~\ref{sec: adaptive_navigation_scheme}, Fig.~\ref{fig:overview_states_localization}(h)), and determine which of the next VO modalities to use:
	\item[2.2.a] Extract planes (Sec.~\ref{sec: plane extraction}) from dense pointclouds (Sec.~\ref{sec: depth map computation}), filtered using the probabilistic map computed in the first stage to prevent large drifts and noise artifacts. See Fig.~\ref{fig:overview_states_localization}(g)and Fig.~\ref{fig:second_stage_pc_comet}.
	\item[2.2.b] Extract and track robust 2D features from imagery; see Sec.~\ref{sec: feature-based method}, Fig.~\ref{fig:overview_states_localization}(f).
	\item[2.3.] Compute visual odometry either from plane registration or feature tracking (Sec.~\ref{sec: plane-based method},~\ref{sec: feature-based method}) depending on the image quality assessment (IQA) and integrate the results into the localization filter.
	%\item[2.4.] Integrate these odometry values into an Extended Kalman filter. Given the mulitmodal inputs  (Fig.~\ref{fig:overview_states_localization} -- baseline and extended input), a mechanism is introduced that validates the reliability of inputs and accordingly adapts inputs for the Kalman filter in order to achieve an enhance localization regarding computational costs and accuracy; see Sec.~\ref{sec: adaptive_navigation_scheme}, Fig.~\ref{fig:overview_states_localization}(h).  
\end{description}

The objective of the first stage is to compute a probabilistic map (octomap~\cite{Hornung2013}) of the expected ROV workspace area.
A coarse 3D representation of the scene can be obtained given few samples.
%Fig.~\ref{fig:first_stage_simulated_octomap} and~\ref{fig:first_stage_real_octomap} 
Figure~\ref{fig:overview_states_localization}(c)
illustrates this by comparing the octomap generated with a simulated RGB-D camera (reference) and the one generated by pointclouds from stereo imagery.
High precision is not crucial here since mapping is not the final goal, but to filter spurious 3D pointclouds, e.g., from dynamic objects like fish or from processing artifacts, as shown in Fig.~\ref{fig:second_stage_pc_comet}.
We will now describe in detail each component from the second stage, \emph{optimized localization}. 

\subsection{Knowledge-enabled localization}
\label{sec: knowledge-enable localization}

Underwater missions are cost-intensive and high-risk, thus prior knowledge about the mission reduces the risk of failures and increases safety. 
Especially in visual inspection or manipulation tasks of man-made structures, the incorporation of prior knowledge can be exploited. 
Therefore, we built a \emph{knowledge base} that contains properties of task-related objects. 
Along with offline information, like CAD models and kinematic descriptions of the robot and OGP panel, the knowledge base is updated based on current information gathered over the execution of a task, e.g.\ panel valve poses.

\subsubsection{Panel Pose Estimation}
\label{sec:method:panel_det}
The panel pose estimation is the basis for projecting the panel model and its kinematic properties into the world-model.
This further enables reliable task benchmarking in simulation and real operations, i.e., manipulation of valves and handles.
Our approach incorporates offline knowledge such as the panel CAD model and visual markers placed at predefined locations, see Fig.~\ref{fig:overview_states_localization}(a).
Based on this augmentation of the panel with markers, we exploit the panel as a fixed landmark and infer the robot pose whenever a visual marker is in the camera view as described in the next Sec.~\ref{sec:marker localization}.	
The panel pose in odometry frame \trfFr{\framePanel}{\frameOdom} can be reliably estimated using the detected marker pose w.r.t.~the camera frame \trfFr{\frameMarker}{\frameCamera}, the camera pose on the robot frame \trfFr{\frameCamera}{\frameRobot}, the panel pose in marker frame \trfFr{\framePanel}{\frameMarker}, and the current robot pose in odometry frame \trfFr{\frameRobot}{\frameOdom}, see Fig.~\ref{fig:overview_states_localization}(e):%~\ref{fig:rov_panel_tf}:
\begin{equation}
	\trfFr{\framePanel}{\frameOdom} = \trfFr{\frameRobot}{\frameOdom} \trfFr{\frameCamera}{\frameRobot} \trfFr{\frameMarker}{\frameCamera} \trfFr{\framePanel}{\frameMarker}
\end{equation}
%\begin{figure}[tb]
%	\small
%	\centering
%	
%	\subfigure[Robot--panel space transformations]{\label{fig:rov_panel_tf}\includegraphics[width=0.65\linewidth]{fig/localization/localization_transforms.pdf}}
%	\subfigure[RGB camera image]{\label{fig:sim_loop_panel_det_real}\includegraphics[width=0.32\linewidth]{fig/localization/real_marker_scaled.png}}	
%	%	\subfigure[Panel with projected kinematic model]{\label{fig:sim_loop_panel_det_km}\includegraphics[width=0.7\linewidth]{fig/localization/rviz_tf.png}}
%	\caption{\subref{fig:rov_panel_tf} Space transformations between robot and oil\&gas panel. \subref{fig:sim_loop_panel_det_real} RGB image of panel with detected markers.}
%	\label{fig:sim_loop_panel_det}
%\end{figure}

%Consequently, $n$ marker observations lead to $n$ panel pose estimates $\trfFr{\framePanel}{\frameOdom}$ that eventually allow to compute the pose mean (see Fig~\ref{fig:sim_loop_panel_det_km}) which includes mean position and orientation, determined by \emph{spherical linear interpolation (Slerp)}~\cite{Shoemake1985}.
When $n$ different markers are constantly detected during $k$ image frames $I$, $n$ pose estimates $\trfFr{\framePanel}{\frameOdom}$ are extracted. 
These are used to compute the mean position and orientation, determined by \emph{spherical linear interpolation (Slerp)}.%~\cite{Shoemake1985}. 

\subsubsection{Visual Landmark-Based Odometry}
\label{sec:marker localization}
%Accurate self-localization of vehicles is a challenging task, especially in the deep-sea domain, due to noisy sensor readings typically based on acoustic devices like Ultra-Short Baseline (USBL) systems, single-beam or multi-beam sonars, Doppler Velocity Log (DVL) or relative readings provided by Inertial Navigation Systems (INS).
%Consequently, localization methods rely on multiple modalities to increase reliability~\cite{ChengSurvey2013,Li2016} and acccal and well-established approach to deal with sensor fusion is the Extended Kalman filter (EKF) \cite{Moore2014} which allows to incorporate these modalities while considering their individual uncertainty.

%However, as discussed in the previous section, reliable dexterous manipulation is a requirement in the DexROV scenario.
%To ensure robust control of the manipulator arm, accurate robot pose estimates are needed.
%Hence, we exploit the panel as a visual landmark again due to its static pose on the seafloor and its visual augmentation with multiple markers.
Once the panel pose has been estimated and fixed, the robot pose can be inferred every time there is a visual marker observation and used as an Extended Kalman Filter input modality.% (Fig.~\ref{fig:overview_states_localization}(e)). 

%In the following we describe our EKF-based localization system incorporating sensor readings and visual landmarks.
%However, one of the main tasks in the DexROV scenario is accurate, dexterous manipulation of valves and levers on man-made structures.
%Therefore, accurate robot pose estimates are a crucial requirement to guarantee reliable manipulation.
%Due to the static position of the panel on the seafloor which is augmented with multiple visual markers, we exploit the panel as a visual landmark.
%Based on its pose estimates, the robot pose can be inferred and used as additional EKF input modality. 
%In the following we describe our EKF-based localization system incorporating sensor readings and visual landmarks.

%\subsubsection{Sensor Readings}
%The robot setup provides a bank of sensors including INS, DVL, and USBL which together allow to gather readings  of the current robot state regarding translation, orientation, linear/angular velocities and accelerations.

%Fig.~\ref{fig:sim_loop_panel_det_real} 
Fig.~\ref{fig:overview_states_localization}(e) shows a sample pose estimate of a visual marker; note that the panel is partially observed but the marker is used to infer the panel pose through the space transformations. % shown in Fig.~\ref{fig:rov_panel_tf}.
Since the panel pose is fixed, the robot pose \trfFr{\frameRobot}{\frameOdom} can be estimated as follows: 
\begin{equation}
\trfFr{\frameRobot}{\frameOdom} = \trfFr{\framePanel}{\frameOdom} \trfFr{\frameMarker}{\framePanel} \trfFr{\frameCamera}{\frameMarker} \trfFr{\frameRobot}{\frameCamera}
\end{equation}
where \trfFr{\framePanel}{\frameOdom} is the panel pose in odometry frame, \trfFr{\frameMarker}{\framePanel} is one marker pose in panel frame, \trfFr{\frameCamera}{\frameMarker} is the camera pose w.r.t.\ the marker and \trfFr{\frameRobot}{\frameCamera} is the robot fixed pose w.r.t.\ the camera.
Further on, the means of robot position $\posFrMean{\frameRobot}{\frameOdom}$ and orientation $\quatFrMean{\frameRobot}{\frameOdom}$ w.r.t.\ the odometry frame are estimated from multiple marker detections using \emph{Slerp}.
In addition, a covariance matrix \covFr{\frameRobot}{\frameOdom} for the robot pose is computed: 
%In addition to the robot pose, the localization component requires a covariance matrix \covFr{\frameRobot}{\frameOdom} which looks like follows:
\begin{equation}
\covFr{\frameRobot}{\frameOdom}=\mathrm{diag}(\sigma^2_{\pos_{x}},\sigma^2_{\pos_{y}},\sigma^2_{\pos_{z}},\sigma^2_{\quat{q}_{\phi}},\sigma^2_{\quat{q}_{\theta}},\sigma^2_{\quat{q}_{\psi}}).
\end{equation}

The full robot pose estimate $\trfFr{\frameRobot}{\frameOdom} = \langle \posFrMean{\frameRobot}{\frameOdom},\quatFrMean{\frameRobot}{\frameOdom} \rangle$ along with the respective covariance matrix \covFr{\frameRobot}{\frameOdom} is then taken as an input for the localization filter in the final setup. 
Alternatively, it can be used as a ground truth value to optimize parameters, i.e., sensor biases and associated covariances, since it is difficult to acquire absolute global ground truth underwater. For more details refer to our work in~\cite{Mueller2018_DexROVSIL}.  

%\subsubsection{Extended Kalman Filter}
%In this work, we apply an \emph{Extended Kalman Filter} (EKF) \cite{Moore2014} to estimate the robot pose over time considering a state space consisting of position $x, y, z$, orientation $\phi, \theta, \psi$, translational  $\dot{x}, \dot{y}, \dot{z}$, and angular velocities $\dot{\phi}, \dot{\theta}, \dot{\psi}$ as well as translational accelerations $\ddot{x}, \ddot{y}, \ddot{z}$.
%We only incorporate direct sensor measurements to the EKF, no integrated or differentiated values. INSs produce angular and linear accelerations, a DVL provides position outputs in form of altitude readings and linear velocities, and the mentioned landmarks are incorporated as pose readings. To increase the localization filter robustness, obvious outliers from sensor readings are rejected heuristically, and the pose inputs inferred from visual markers are tuned based on our experimental results.

\subsection{Dense depth mapping and plane extraction}
\label{sec: dense depth mapping and computation}

\subsubsection{Depth map computation}
\label{sec: depth map computation}
This is a preprocessing step for the plane-based odometry computation (Sec.~\ref{sec: plane-based method}).
%The pipeline has a number of preprocessing steps, one of which is computing a dense depth map from stereo images.
We consider two state-of-the-art real-time dense point cloud (DPC) generators: Efficient Large-Scale Stereo Matching (ELAS)~\cite{Geiger2011_ELAS} and Dispnet~\cite{Mayer2016_Dispnet}.
ELAS is geometry based and relies on a probabilistic model built from sparse keypoints matches and a triangulated mesh, which is used to compute remaining ambiguous disparities.
Since the probabilistic prior is piece wise linear, the method is robust against uniform textured areas.
Dispnet is a data driven approach that uses a convolutional neural network for disparity regression.

The reason to compare these methods is due to their underlying distinct approaches, which in turn offer different advantages and disadvantages. 
For example, ELAS has faster processing times in CPU and outputs more precise but incomplete maps when specular reflections or very large textureless areas occur.
Dispnet produces a complete map even in the presence of image distortions, but smooths the disparity changes around object boundaries. 

On top of the depth estimation, it is important to include techniques that reduce noise-induced artifacts commonly present when performing outdoor stereo reconstruction. 
Fig.~\ref{fig:second_stage_pc_comet} shows an example where object borders produce comet-like streaks fading away from the camera position.
We also encountered this artifact when running the system with Dispnet during sea field trials.
It was observed that when the GPU RAM memory gets overloaded some layers of the Dispnet neural network produce spurious results.
%This is the result of many systems running in the ROV during deep-sea trials besides the localization filter i.e., valve detection/manipulation, camera-feed transmission, GUI visualizations, etc. 

In order to mitigate these artifacts, the incoming point cloud is filtered by rejecting points which do not intersect with the octomap~\cite{Hornung2013} representing the workspace obtained from the first navigation stage (\emph{loose localization}).
For an efficient nearest-neighbor search between point cloud and octomap voxels, a \emph{kd}-tree is applied to represent the workspace octomap.
Consequently, a substantial amount of points is neglected (Fig.~\ref{fig:overview_states_localization}(g) -- depth map point cloud generation) and also reduces computation cost and memory.

\subsubsection{Plane extraction}
\label{sec: plane extraction}
Due to the noisy nature of stereo generated pointclouds, we used a region growing based technique for plane extraction \cite{poppinga2008fast}.
It also outputs a covariance matrix that describes the planarity of the found planes, which can then be integrated for a better estimation of the localization uncertainty (Fig.~\ref{fig:overview_states_localization}(g) -- plane extraction-based visual odometry).
Moreover, it efficiently represents not only planes but found holes as polygons, which allows to reason about the quality of the data. 

In summary, point clouds are segmented into several planes $\mathbb{\varPi} = \{\pi_i | i = 1,...,N\} $. 
Initially, an arbitrary point $p_0$ and its nearest neighbor $p_1$ form an initial set $\mathbb{P}_i$ representing the plane $\pi_i$. 
Then the set $\mathbb{P}_i$ grows by finding its nearest neighbors $p_j$ and adding $p_j$ to $\mathbb{P}_i$ if $p_j$ belongs to $\pi_i$ (plane test), and stops when no more points can be added. 
Afterwards, the process continues until all the points are distributed into a plane set $\mathbb{P}$ or considered as noise.
 
\subsection{Visual odometry}
\label{sec:visual odometry}

The visual markers attached to the panel (Sec.~\ref{sec:marker localization}) are not always observable. 
Therefore, further methods are beneficial to aid navigation. 
Here we adapt plane-based and featured-based odometry methods to our scenario to exploit structures and features found in the environment.
%where the former can match planes quickly and the latter provides accurate pose given enough features in the environment.

\subsubsection{Odometry from plane registration}
\label{sec: plane-based method}

After plane segmentation (Sec.~\ref{sec: plane extraction}), the plane normal equals to the eigen vector with smallest eigen value of Matrix $\mathbf{A}$:
\begin{equation}
\small
	\mathbf{A} = \begin{pmatrix}
	\Gamma_n(x,x) & \Gamma_n(x,y) & \Gamma_n(x,z)\\
	\Gamma_n(y,x) & \Gamma_n(y,y) & \Gamma_n(y,z)\\
	\Gamma_n(z,x) & \Gamma_n(z,y) & \Gamma_n(z,z)
	\end{pmatrix}
\end{equation}

where $\Gamma_n(\alpha,\beta) = \sum_{j}^{n}(\alpha_j - m_{\alpha})(\beta_j - m_{\beta}), \alpha, \beta \in \{x, y, z\}$ and $m$ is the mass center of the points in plane set $\mathbb{P}_i$. 
To update the matrix $\mathbf{A}$ and hence the planes normal representation as fast as possible when new points are considered, the sum of orthogonal distances $\Gamma_{l}(\alpha,\beta)$ is updated with a new point $p_{l+1}$ as follows:

\begin{equation}
\small
\begin{aligned}
\Gamma_{l+1}(\alpha,\beta) = & \Gamma_l(\alpha,\beta) + \alpha_{l+1}\beta_{l+1} \\
& -m_{\alpha}(l+1)(\sum_{j=1}^{l+1}p_j + m_{\alpha}(l+1)) \\
& +m_{\alpha}(l)(\sum_{j=1}^{l}p_j + m_{\alpha}(l))
\end{aligned}
\end{equation}

Then the relative pose ${^C_R\mathbf{T}_{rel}}$ in camera frame at time $t$ and $t+1$ can be calculated by the extracted planes. 
Here we exploit the plane registration method in \cite{pathak2010fast} to estimate rotation only.
As shown in Sec.~\ref{exp: dense maps} experiment, in our deep-sea scenario we commonly encountered between 3 to 5 planes per frame, and at least 4 plane correspondences are necessary to estimate translation.  

Suppose planes extracted at time $t$ and $t+1$ be $\varPi_1  = \{{^1\pi_i} | i = 1,...,M\} $ and $\varPi_2  = \{{^2\pi_j} | j = 1,...,N\} $ respectively, then the $M \times N$ candidate pairs $({^1\pi_i}, {^2\pi_j})$ are filtered and registered by the following tests from~\cite{pathak2010fast}, which are adapted to our deep-sea setup:

\begin{itemize}
	\item Size-Similarity Test: 
	The Hessian matrix $H$ of the plane parameters derived from plane extraction is proportional to the number of points in the plane $\pi$. Thus, the Hessian of two matched planes should be similar, i.e., 
	\begin{equation}
		\small
		\lvert log{\lvert ^1H_i \rvert_+} - log{\lvert ^2H_i \rvert_+} \rvert < \bar{L}_{det}
	\end{equation}
	where $\lvert H_i \rvert_+$ is the product of the singular values of $H_i$ and $\bar{L}_{det}$ is the similarity threshold.
	 
	\item Cross-Angle Test: The angle between two planes $({^1\pi_{i_1}}, {^1\pi_{i_2}})$ in frame ${^1\mathcal{F}}$ should be approximately the same as the angle to the correspondent two planes $({^2\pi_{j_1}}, {^2\pi_{j_2}})$ in frame ${^2\mathcal{F}}$, described as
	\begin{equation}
		\small
		{^1\hat{n}^{\top}_{i_1}}{^1\hat{n}_{i_2}} \approx {^2\hat{n}^{\top}_{j_1}}{^2\hat{n}_{j_2}}
	\end{equation}
	where $\hat{n}_{k}$ is the normal to the plane $\pi_k$.
	
	\item Parallel Consistency Test: Two plane pairs $({^1\pi_{i_1}}, {^2\pi_{j_1}})$ from the frames ${^1\mathcal{F}}$ and ${^2\mathcal{F}}$ are defined as parallel if their normals meet ${^1\hat{n}^{\top}_{i_1}}{^1\hat{n}_{i_2}}\approx 1$ and ${^2\hat{n}^{\top}_{j_1}}{^2\hat{n}_{j_2}} \approx 1$, or ${^1\hat{n}^{\top}_{i_1}}{^1\hat{n}_{i_2}}\approx -1$ and ${^2\hat{n}^{\top}_{j_1}}{^2\hat{n}_{j_2}} \approx -1$.
\end{itemize}

If only one plane is extracted from the current frame, it is tested only by the Size-Similarity test because others require at least two plane correspondences. 
Then the filtered plane pairs are used to calculate the rotation ${^1_2R}$ between frame ${^1\mathcal{F}}$ and ${^2\mathcal{F}}$ by the equation: 
\begin{equation}
\small
\max_{^1_2R} \zeta_r = constant + \sum_{i=1}^{S}\omega_i{^1\hat{n}_i}\cdot({^1_2R} {^2\hat{n}_i})
\label{eq: cal_rotation}
\end{equation}

which can be solved by Davenport's q-method and where $w_i$ are weights inversely proportional to the rotational uncertainty. 
If the rotation ${^1_2R}$ is represented as quaternion ${^1_2\hat{q}}$, Eq.~\ref{eq: cal_rotation} can be written as:
\begin{equation}
\small
\begin{aligned}
\max_{^1_2\hat{q}} \zeta_r & = \sum_{i=1}^{S}\omega_i{^1_2\hat{q}}K {^1_2\hat{q}}
\end{aligned}
\end{equation}

Then the covariance ${^1_2\mathbf{C}_{\hat{q}\hat{q}}}$ of quaternion ${^1_2\hat{q}}$ is 
\begin{subequations}
	\small
	\begin{equation}
	\small
	{^1_2\mathbf{H}_{\hat{q}\hat{q}}} = 2(K - \mu_{max}(K)I)
	\end{equation}
	\begin{equation}
	\small
	{^1_2\mathbf{C}_{\hat{q}\hat{q}}} = - {^1_2\mathbf{H}_{\hat{q}\hat{q}}}^{+}
	\end{equation}
\end{subequations}

where $\mu_{max}(K)$ is the maximum eigen value of $K$, derived from Davenport's q-method. The covariance ${^1_2\mathbf{C}_{\hat{q}\hat{q}}}$ and rotation ${^1_2\hat{q}}$ are used as input for the our navigation filter. 

%For each corresponding planes $(^1\pi_i, ^2\pi_i)$, the transformation can be described as
%\begin{subequations}
%	\begin{equation}
%	{^1\hat{n}_i} = {^1_2R} {^2\hat{n}_i}
%	\end{equation}
%	\begin{equation}
%	{^1\hat{n}_i}{^1_2t} = {^1d_i} - {^2d_i}
%	\end{equation}
%\end{subequations}
%where $\hat{n}\cdot\eta = d$ is the plane equation.
%When there are more than one plane pairs, the rotation ${^1_2R}$ can be found by maximizing the following equation
%\begin{equation}
%\begin{aligned}
%	\zeta_r & = -\frac{1}{2}\sum_{i=1}^{M}\omega_i\lVert s \rVert^2 \\
%	& = constant + \sum_{i=1}^{M}\omega_i{^1\hat{n}_i}({^1_2R}{^2\hat{n}_i})
%\end{aligned}
%\end{equation}
%which can be solved by Davenport's q-method. Then the difficulty is to find the corresponding planes. As described in \cite{pathak2010fast}, we found all the possible pairs between these two plane sets and then filtering the pairs by other tests. The most important condition is that the plane pairs should meet the cross-angle test, shown as
%\begin{equation}
%	\chi^2_x = \frac{({^1\hat{n}^{\top}_{j_1}}{^1\hat{n}_{i_1}} - {^2\hat{n}^{\top}_{j_2}}{^2\hat{n}_{i_2}})^2}{{^1\sigma^2_{i_1,j_1}} + {^2\sigma^2_{i_2,j_2}}}
%\end{equation} 
%where $^1\sigma^2_{i_1,j_1} \approx {^1\hat{n}^{\top}_{j_1}}{^1D_{i_1}}{^1\hat{n}_{j_1}} + {^1\hat{n}^{\top}_{i_1}}{^1D_{j_1}}{^1\hat{n}_{i_1}}$, ${^1D_{i_1}}$ is the covariance matrix of ${^1\hat{n}_{i_1}}$ and ${^2\sigma^2_{i_2,j_2}}$ is similar.

\subsubsection{Feature-based tracking}
\label{sec: feature-based method} 
Whenever there are distinctive and sufficient 2D texture features in the environment, related methods provide a reliable and fast way to compute odometry. 
Here, ORB-SLAM2 \cite{mur2017orb} is used. It consists of three main threads: tracking, local mapping, and loop closing. 
Considering our application, we briefly describe the tracking process. 
When a new stereo image is input, the initial guess of the pose ${^C_R\mathbf{T}'}$ is estimated through the tracked features from the last received image. 
Afterwards, the pose ${^C_R\mathbf{T}}$ can be improved by conducting bundle adjustment on a memorized local map $\mathbf{M}_i$. 
Moreover, the tracking thread also decides whether the stereo image should be an image keyframe or not. 
When tracking fails, new images are matched with stored keyframes to re-localize the robot.

ORB-SLAM2 was chosen because direct VO methods (DSO,LSD-SLAM) assume brightness constancy throughout image regions~\cite{Park2017_slam}, which seldom happens in underwater due to light backscatter.
Likewsie, visual-inertial SLAM methods (VINS,OKVIS) require precise synchronization between camera and IMU~\cite{Buyval2016_slam}, but by hardware design all sensors are loose-coupled in our application.

\subsection{Adaptive image quality based navigation}
\label{sec: adaptive_navigation_scheme}

%Problems: computation processing, localization accuracy of particular cues degrades when sensory quality decrease. 
%Need to compensate and identify a decision criteria when to incorporate which modalities (when is it expected to cue provides a good result.) 
%We have introduced multiple modalities see Fig -> baseline local and extend.
%Baseline focus on on-board sensor bench which provides (relative) localization estimates. 
%How to decide -> image quality IQA

At the end of the localization pipeline the EKF can integrate all the inputs based on their measurement confidence, i.e., covariance matrix.
For efficiency, it is preferable to filter out low confidence odometry values before using them for the EKF. 
This could be done by examining the covariance matrix after the vision processing.
But computation time is an important factor  in our real-time application. We hence use decision criteria on the sensor (image) quality to omit visual odometry computations, which are likely to generate low confidence results. 

We introduced multiple visual odometry modalities in Sec.~\ref{sec:visual odometry}; see Fig.~\ref{fig:overview_states_localization}(e)(f)(g).
The visual marker-based odometry, as part of the baseline inputs, is not filtered out due to its high reliability and precision.
Feature tracking ORB-SLAM localization is highly dependent on image quality; textureless regions and low-contrast significantly reduce its accuracy.
In contrast, plane-based odometry copes well with textureless environments given that there is an underlying structure. 
But it is very computationally demanding due to dense depth estimation and plane extraction (Sec.~\ref{sec: dense depth mapping and computation}).

Based on this, we propose an image quality assessment (IQA) to reason about which visual cues to use in the localization pipeline.
We aggregate a non-reference image quality measure based on Minkowski Distance (MDM)~\cite{Ziaei2018_MDMIQA} and the number of tracked ORB features between consecutive frames.
The MDM provides three values in the $\left[ 0,1 \right]$ range describing the contrast distortion in the image; thus, the number of ORB features is normalized based on the predefined maximum number of features to track. 
If each of these IQA values is defined as $m_I(t)$, the final measurement for each timestamp $t$ is their average $\overline{m}_I(t)$.
Experiments in Sec.~\ref{exp:visual odometry performance} show the performance of these IQA measurements.

%
%%=== EXPERIMENTS ===%
\section{Experimental Results}
\label{sec:experiments}

%To evaluate our proposed methodology, we conduct three experiments with  different objectives.
The first two experiments are performed in simulation to analyze their algorithmic behavior.
%, i.e., excluding noise sources such as motion blur, dynamic lighting conditions, specular surfaces, among others.
The simulator ambient lighting parameters are adjusted to match conditions from the sea trials.
%Like in our previous work~\cite{Mueller2018_DexROVSIL}, this was performed using the FSIM image quality metric. 
%Like in other work~\cite{Mueller2018_DexROVSIL}, this was performed using the FSIM image quality metric. 
The simulated ROV navigates around while keeping $\approx$\SI{1.5}{m} from the OGP panel since it was found to be an optimal distance for our stereo camera baseline of \SI{30}{cm}; also a constant $z$-axis value (depth) is kept.

First, we evaluate the impact of the dense pointcloud generators, ELAS and Dispnet, on the plane extraction, registration and orientation computation (Sec.~\ref{sec: plane extraction},~\ref{sec: plane-based method}).
Furthermore, we study how the filter based on the probabilistic map generated from the \emph{first stage} of our navigation scheme improves performance.
The second experiment assesses the accuracy of the VO approaches, i.e., our plane registration and feature tracking (ORB-SLAM2), with different types of imagery.
%
%These two first experiments are performed in simulation in order to find their algorithmic behavior; in other words, excluding noise sources such as motion blur, dynamic lighting conditions, specular surfaces, among others.
%However, the simulator static and background ambient lighting parameters are heuristically adjusted to resemble captured data from field trials.
%Like in our previous work~\cite{Mueller2018_DexROVSIL}, this was performed using the FSIM image quality metric. 
%Like in other work~\cite{Mueller2018_DexROVSIL}, this was performed using the FSIM image quality metric. 
%The simulated ROV navigates around the panel keeping approximately \SI{1.5}{m} distance from the oil\&gas panel since it was found to be an optimal distance for the used stereo camera baseline of \SI{30}{cm}; also a constant $z$-axis value (depth) is kept.
%
The last experiment tests our complete pipeline with real-world data from DexROV field trials in the sea of Marseille, France (see Fig.~\ref{fig:dexrov_in_action}).
The cameras used are Point Grey Grasshoppers2 which have a resolution of $688 \times 516$ pixels and \SI{10}{Hz} rate; both are in underwater housings with flat-panel that allows for image rectification using the PinAx model~\cite{Luczynski2017_pinax}.      
 
\subsection{Plane segmentation from dense depth maps}
\label{exp: dense maps}

In this first experiment, we investigate how the plane extraction and registration algorithms perform with different dense point cloud generators.
%As mentioned, this is performed in simulation to study the algorithmic expected error without concerning about real-world error sources i.e., camera distortions, lost of sensor inputs, hardware malfunctions, etc.  
%
Table~\ref{tab:depth_map_planes} shows the experiment results. 
We use as a baseline the simulated RGB-D camera available in the Gazebo simulation engine, which provides ground truth depth/disparity maps.
To measure the accuracy of the stereo disparities (second column) the same principle as the 2015 KITTI stereo benchmark was followed~\cite{Menze2015_sceneflow}, all disparity differences greater than 3 pixels or $5\%$ from the ground truth are considered erroneous. 
The coverage score (third column) counts how many image pixels have a valid associated disparity value; textureless regions reduce this value.
Furthermore, we also count the number of extracted planes and holes within them (fourth and fifth column) using the method from Sec.~\ref{sec: plane extraction}. 
Finally, the last column of Table~\ref{tab:depth_map_planes} shows the orientation error computed from the plane registration, Sec.~\ref{sec: plane-based method}.

\begin{table}[!b]
	\footnotesize
	\centering
	\captionsetup{justification=centering}
	\caption{Dense map, plane extraction and orientation measures on simulated stereo sequences}
	\begin{tabularx}{\linewidth}{@{}XYYzzY@{}}
		\toprule
		\textbf{Method} & \textbf{Accuracy} & \textbf{Coverage} & \textbf{Planes} & \textbf{Holes} 
		& \textbf{Error} $[^\circ]$ \\
		\midrule
		RGB-D 			& 1.0 	& 1.0 	& 1620 & 456 & $11.7\pm3.3$ \\
		ELAS 			& 0.781 & 0.579 & 2708 & 713 & $16.2\pm7.3$ \\
		Dispnet 		& 0.713 & 0.943 & 1987 & 123 & $19.4\pm8.5$ \\
		ELAS+Filter 	& 0.854 & 0.468	& 2061 & 204 & $12.1\pm6.4$ \\
		Dispnet+Filter 	& 0.798	& 0.833	& 1254 & 18	 & $09.3\pm2.1$ \\
		\bottomrule
	\end{tabularx}
	\label{tab:depth_map_planes}
\end{table}

It is important to note that the tests are performed one more time using the probabilistic map generated from the \emph{first stage} of our methodology (Fig.~\ref{fig:overview_states_localization}(c)) as a filter.
We draw the next conclusions from this experiment: 

\begin{enumerate}
	\item[1)] ELAS depth maps have more accurate 3D information at the cost of incomplete maps, which produce higher number of planes and holes due to the inability to connect regions corresponding to the same planes.
	\item[2)] Since these redundant planes are still accurate space representations, they produce better orientation estimations than the complete but inaccurate Dispnet maps.
	\item[3)] Filtering point clouds with the probabilistic map boosts accuracy and reduces the coverage of both methods, which validates the efficiency of our two-stage navigation scheme. 
	%\item[4)] The probabilistic filtering makes the plane-based orientation computation have less error in each case, which validates the efficiency of our two-stage navigation scheme.
	\item[4)] Dispnet$+$Filter orientation accuracy is even greater than the one based on the simulated RGB-D camera. From our observations, Dispnet$+$Filter generates less planes than the RGB-D depth maps; RGB-D very high accuracy produces planes for small objects such as the panel's valves which add ambiguity.
	\item[5)] Thus, highly accurate point clouds (overfitting) negatively affects plane registration, i.e., the likelihood of incorrectly registering nearby small planes increases.
	\item[6)] Based on the 367 analyzed image frames, the mean number of planes generated with Dispet+Filter per frame is $3$ or $4$.      
\end{enumerate}   

\subsection{Image quality based navigation performance}
\label{exp:visual odometry performance}

\begin{figure}[!b]
	\centering
	\subfigure[Side 1]{\label{fig:panel_side_1}
		\includegraphics[width=0.3\linewidth]{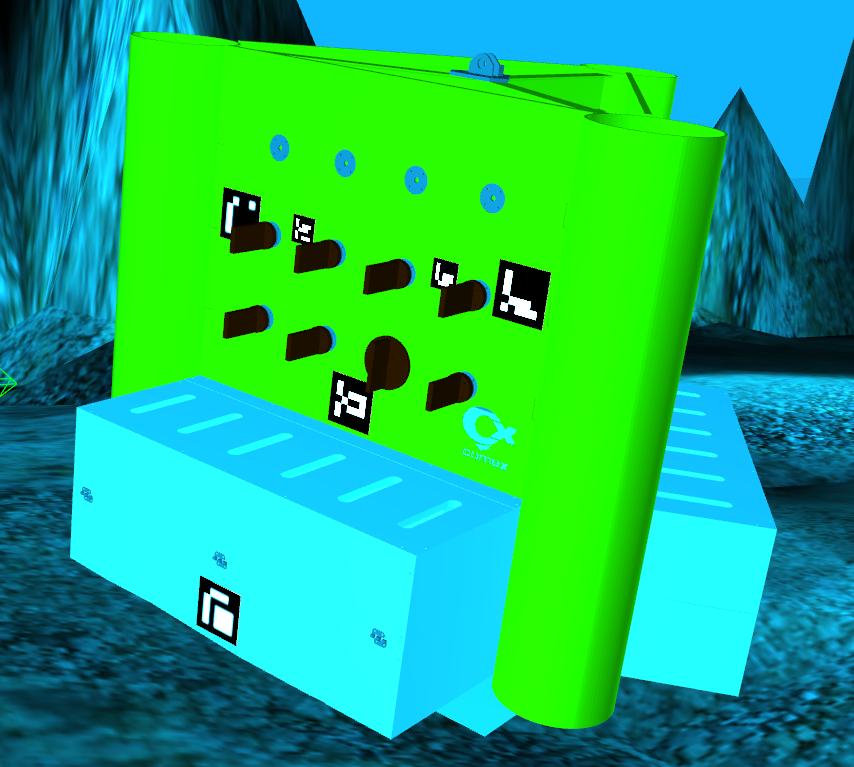}}
	\subfigure[Side 2]{\label{fig:panel_side_2}
		\includegraphics[width=0.3\linewidth]{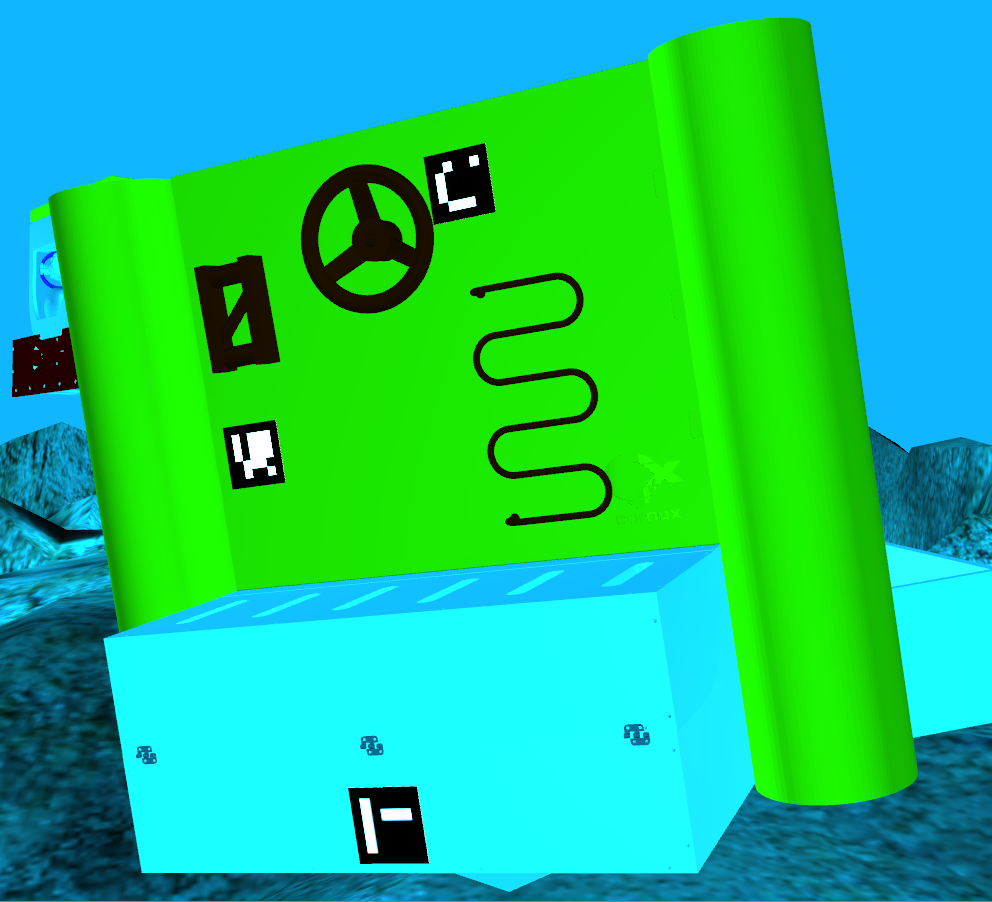}}
	\subfigure[Side 3]{\label{fig:panel_side_3}
		\includegraphics[width=0.33\linewidth]{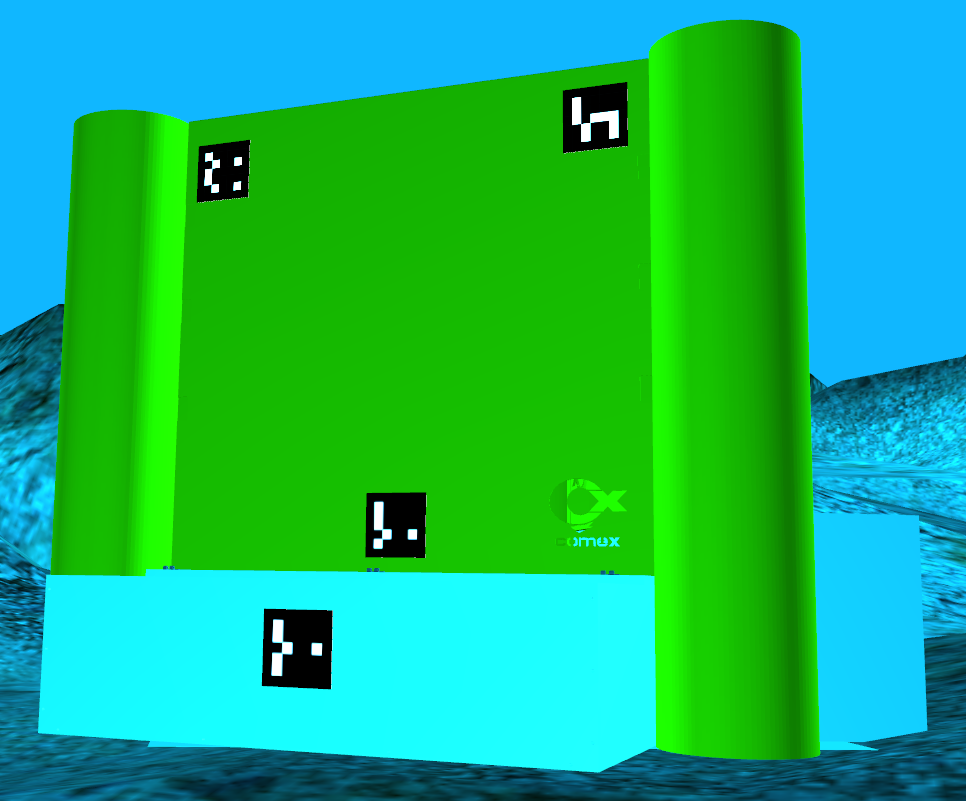}}
	\caption{Oil\&gas panel sides for \subref{fig:panel_side_1} valve manipulation, \subref{fig:panel_side_2} dexterous grasping, \subref{fig:panel_side_3} and textureless region visualization.}
	\label{fig:panel_in_simulation}
\end{figure}

Based on the previous experiment, we choose Dispnet as dense point cloud generator.
To analyze the strengths and weaknesses of the VO methods, the ROV circles the panel while acquiring very diverse stereo imagery.
The panel has three sides with distinct purposes: valve manipulation (side 1), dextereous grasping (side 2) and textureless side (side 3), see Fig.~\ref{fig:panel_in_simulation}.  
%Nonetheless, in simulation the side with biological samples (corals, rocks, etc) has not been modeled due to the complexity of the required objects. 
This last panel side helps evaluating how the methods work with scarce image features, and how our image quality measure $\overline{m}_I(t)$ from Sec.~\ref{sec: adaptive_navigation_scheme} evaluates this.

%\begin{figure}[!b]
%	\centering
%	\resizebox{\linewidth}{!}{\inputpgf{fig/image_quality}{exp-4-2_.pgf}}
%	\caption{(Top) Orientation error for different visual odometry methods. No markers detected for sampling times marked \textcolor{orange}{orange}, and changes of panel side with a \textcolor{red}{red} line. (Bottom) Image quality measurement $\overline{m}_I(t)$ per stereo pair.}
%	\label{fig:visual odometry simulation tests}
%\end{figure}

The expected error pose is defined as the difference between the ground-truth robot pose in simulation $\robotPoseSimulation$ and the robot pose determined from visual odometry $\robotPoseVO$:

\begin{equation}
\small
\begin{aligned} 
\small
m(\mathcal{E})= & \robotPoseError{S}{VO} \\
= & \langle \robotPositionError{S}{VO} , \robotOrientationError{S}{VO} \rangle
\end{aligned}
\label{eq:pose_error_measure}
\end{equation}

where $\panelPositionError{S}{VO}$ is the Euclidean distance between positions and $\panelOrientationError{S}{VO}$ is the minimal geodesic distance between orientations.
% \cite{Huynh2009}.
For our experiments, we also compute the \emph{lag-one autocorrelation} $m_{A}=\sum_{t}\robotPoseFilter (t)\robotPoseFilter (t-1)$ on the EKF filter predicted poses $\robotPoseFilter$; $m_{A}$ is a measure of trajectory smoothness, important to prevent the robot from performing sudden jumps.

\subsubsection{Visual odometry accuracy}
\label{exp: vo accuracy}

\begin{figure}[!b]
	\centering
	\resizebox{\linewidth}{!}{\input{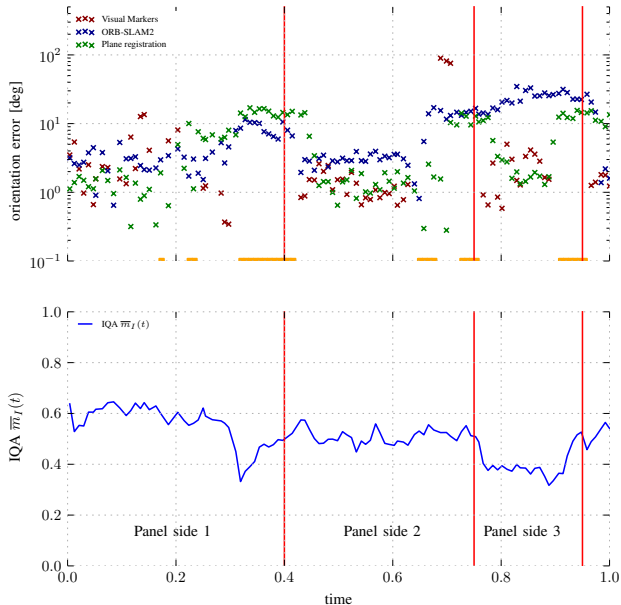}}
	\caption{(Top) Orientation error for different visual odometry methods. No markers detected for sampling times marked \textcolor{orange}{orange}, and changes of panel side with a \textcolor{red}{red} line. (Bottom) Image quality measurement $\overline{m}_I(t)$ per stereo pair.}
	\label{fig:visual odometry simulation tests}
\end{figure}

%For this experiment, all poses are computed in the odometry frame $O$ (see Fig.~\ref{fig:rov_panel_tf}); but we only use orientation because the number of encountered planes per frame is not sufficient to compute translation, as encountered in the previous experiment. 
First, we evaluate the accuracy of the proposed VO methods.
Fig.~\ref{fig:visual odometry simulation tests}(top) shows the results, the time axis has been normalized and the error is logarithmically scaled for better readability.  
The orange horizontal lines indicate the time when no visual marker is detected, and the vertical red lines show when the ROV is transitioning into another side of the panel. 

As expected, the orientation derived from the markers is the most accurate but it also presents the outliers with the largest errors, e.g., close to \SI{0.1}{s} and \SI{0.7}{s} in Fig.~\ref{fig:visual odometry simulation tests}.
The feature tracking ORB-SLAM2 method presents the greatest error but with the least variance. 
When there are very scarce features to track, such as in panel's side 3, the error abruptly increases until a memory-saved keyframe from the panel's side 1 is seen again; close to time \SI{1.0}{s}.
%That is when the panel's side 1 is seen again, close to time \SI{1.0}{s} in Fig.~\ref{fig:visual odometry simulation tests}. 

%\begin{figure}[!b]
%	\centering
%	\resizebox{\linewidth}{!}{\inputpgf{plots/2018}{exp2__orientation_error.pgf}}
%	\caption{Orientation error for different visual odometry methods. No markers detected for sampling times marked \textcolor{orange}{orange}, and changes of panel side with a \textcolor{red}{red} line.}
%	\label{fig:visual odometry simluation tests}
%\end{figure}

%\begin{figure}[!b]
%	\centering
%	\resizebox{\linewidth}{!}{\inputpgf{fig/image_quality}{exp-4-2_.pgf}}
%	\caption{(Top) Orientation error for different visual odometry methods. No markers detected for sampling times marked \textcolor{orange}{orange}, and changes of panel side with a \textcolor{red}{red} line. (Bottom) Image quality measurement $\overline{m}_I(t)$ per stereo pair.}
%	\label{fig:visual odometry simulation tests}
%\end{figure}

%\input{sections/field_trials_localization_plot}

The plane registration method has similar accuracy as the visual marker-based odometry and with scarce outliers only when the ROV transitions between panel sides.  
During these periods the corner of the panel is seen, which is not a planar but cylindrical surface. 
Then, depending on the viewpoint it will be represented with different various planes. 

These results are complemented by Table~\ref{table:localization_all_vs_adaptive}(a), which shows the higher computational costs of the plane-based VO.
During field trials, the overall ROV perception$+$manipulation control system and its graphical interface have reported peaks of $92\%$ GPU RAM usage; hence, using Dispnet can lead to GPU overuse and spurious 3D maps. 
Likewise, the slow update times of plane-based VO might limit the ROV velocity.
For these reasons, VO based on feature tracking is given preference when the image quality is good. 
%Regardless, it is preferable to have another input modality at all times in addition to navigational sensors which are prone hardware failures and biases.

\begin{table}[!b]
	\footnotesize
	%\scriptsize
	\captionsetup{justification=centering}
	\caption{Image quality based navigation performance}

	\begin{tabularx}{.45\linewidth}{@{}XYY@{}}
		\toprule
		~ & VO-ORB & VO-planes \\ 
		\midrule
		CPU $\left[ \% \right]$ & 3.2 & 6.8 \\
		GPU $\left[ \% \right]$ & 0.1 & 17.6 \\
		Time $\left[ s \right]$ & 0.145 & 3.151 \\
		\bottomrule
	\end{tabularx}
	\hfill
	\begin{tabularx}{.5\linewidth}{@{}XYY@{}}
		\toprule
		~ & EKF-all & EKF-adpative \\ 
		\midrule
		$\bar{m}_{M,F}(\langle \bar{\mathbf{p}} \rangle )$ $\mathrm{[m]}$ & 0.73 \phantom{a}$\pm$0.38 & 0.61 \phantom{a}$\pm$0.14 \\
		$\bar{m}_{M,F}(\langle \bar{\mathbf{q}} \rangle )$ $\mathrm{[deg]}$ & 8.93 \phantom{a}$\pm$4.22 & 3.02 \phantom{a}$\pm$1.06 \\
		$m_{A}$ & 0.92 & 0.95 \\
		\bottomrule
	\end{tabularx}
	\label{table:localization_all_vs_adaptive}
	\vspace{0.1cm}
	\begin{flushleft}
		(a) Computation performance \hspace{0.7cm}(b) Pose error and traj. autocorrelation
	\end{flushleft}
	
\end{table}

\subsubsection{Image quality assessment}
\label{exp:iqa}

In this experiment, we validate that the proposed image quality measure $\overline{m}_I(t)$ detects when an image has low contrast and/or large uniform texture regions.
This is shown in Fig.~\ref{fig:visual odometry simulation tests}(bottom); 
$\overline{m}_I(t)$ is the lowest when panel side 3 is in view and when the ROV navigates around the corners.
%, i.e., changes from one panel side to another. 
As it can be seen, $\overline{m}_I(t)$ mostly shows an inverse behavior than the VO accuracy with ORB-SLAM2.

Based on this simulations, we set a threshold $(\approx0.45)$ for $\overline{m}_I(t)$ to only trigger the computationally expensive plane-based VO when the image quality is poor.
When using the IQA to decide which VO inputs to integrate into the localization filter (\emph{EKF-adaptive}), we reduce the pose error and increase the smoothness of the followed trajectory, see Table~\ref{table:localization_all_vs_adaptive}(b).
Simply integrating all odometry inputs (\emph{EKF-all}) does not boost performance as the Kalman filter does not reason about the quality of the sensor data except for examining the inputs covariance matrix.

\subsection{Field trials localization}
\label{exp:field trials localization}

%\begin{figure*}[!t]
%	\centering
%	\begin{subfigure}[$\mathcal{T}_{L1}$ - Localization using navigation sensors and visual landmarks]{
%			\label{fig:results_localization_real_markers}
%			\resizebox{0.31\textwidth}{!}{\inputpgf{plots/2018}{bagfile_around_panel_only_markers_trajectory.pgf}}
%		}
%	\end{subfigure}
%	\begin{subfigure}[$\mathcal{T}_{L2}$ - Localization using navigation sensors, visual landmarks and visual odometry]{
%			\label{fig:results_localization_real_markers_planes}
%			\resizebox{0.31\textwidth}{!}{\inputpgf{plots/2018}{bagfile_around_panel_markers_planes_trajectory.pgf}}
%		}
%	\end{subfigure}
%	\begin{subfigure}[$\mathcal{T}_{L3}$ - Localization using navigation sensors and visual odometry]{
%			\label{fig:results_localization_real_plane}
%			\resizebox{0.31\textwidth}{!}{\inputpgf{plots/2018}{bagfile_around_panel_only_planes_trajectory.pgf}}
%		}
%	\end{subfigure}
%	\caption{Robot poses (triangles) with orientation error $\robotOrientationError{S}{F}$ (triangle color) and position error $\robotPositionError{S}{F}$ (circle color and log-scaled circle radius) as the ROV circles the oil \& gas panel. Only the instances where poses from visual markers $\robotPoseMarker$ can be computed are shown since these are used as ground truth.}
%	\label{fig:results_localization_real_trajectory}
%\end{figure*}

\begin{figure*}[!t]
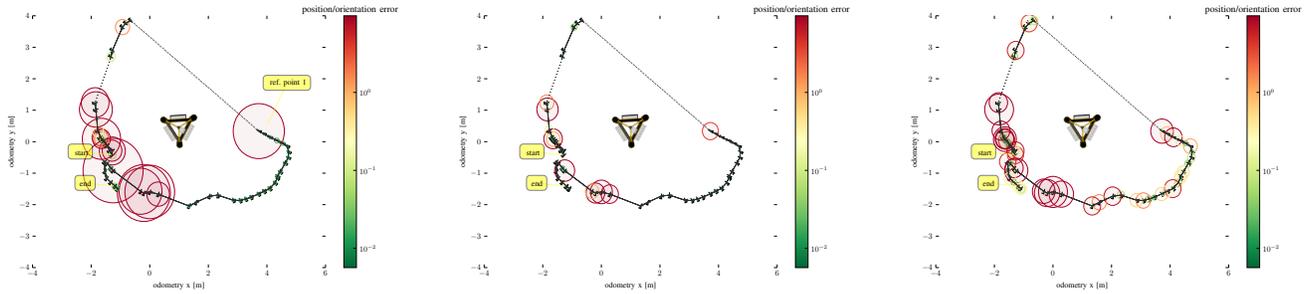

	\centering
	\begin{subfigure}[$\mathcal{T}_{L1}$ - Localization using navigation sensors and visual landmarks]{
			\label{fig:results_localization_real_markers}
			\resizebox{0.31\textwidth}{!}{\input{plots/2018/bagfile_around_panel_only_markers_trajectory.pgf}}
		}
	\end{subfigure}
	\begin{subfigure}[$\mathcal{T}_{L2}$ - Localization using navigation sensors, visual landmarks and visual odometry]{
			\label{fig:results_localization_real_markers_planes}
			\resizebox{0.31\textwidth}{!}{\input{plots/2018/bagfile_around_panel_markers_planes_trajectory.pgf}}
		}
	\end{subfigure}
	\begin{subfigure}[$\mathcal{T}_{L3}$ - Localization using navigation sensors and visual odometry]{
			\label{fig:results_localization_real_plane}
			\resizebox{0.31\textwidth}{!}{\input{plots/2018/bagfile_around_panel_only_planes_trajectory.pgf}}
		}
	\end{subfigure}
	\caption{Robot poses (triangles) with orientation error $\robotOrientationError{S}{F}$ (triangle color) and position error $\robotPositionError{S}{F}$ (circle color and log-scaled circle radius) as the ROV circles the oil \& gas panel. Only the instances where poses from visual markers $\robotPoseMarker$ can be computed are shown since these are used as ground truth.}
	\label{fig:results_localization_real_trajectory}
\end{figure*}

In the following experiments, we use the visual landmarks (markers) pose estimates $\robotPoseMarker$ as ground truth since they are quite accurate and the robot global pose can be obtained from them (Sec.~\ref{sec:marker localization}).
We perform three different tests $\mathcal{T}_{Li}$ explained in Table~\ref{table:localization tests description}; the corresponding results are shown in Table~\ref{table:localization_measures} and Fig.~\ref{fig:results_localization_real_trajectory}.
In these tests we compute the measure $m_{M,F}(\mathcal{T}_{Li})=\robotPoseError{M}{F}$ as defined in equation~\ref{eq:pose_error_measure}, plus the \emph{lag-one autocorrelation} $m_{A}(\mathcal{T}_{Li})$.
%=\sum_{t}\robotPoseFilter (t)\robotPoseFilter (t-1)$ on the EKF filter predicted poses $\robotPoseFilter$; 
%$m_{A}(\mathcal{T}_{Li})$ is a measure of trajectory smoothness, important to prevent the robot from performing sudden jumps.

\begin{table}[!b]
	\centering
	%\scriptsize
	\footnotesize
	%\small
	\captionsetup{justification=centering}
	\caption{Description of localization tests $\mathcal{T}_{Li}$}
	\label{table:localization tests description}
	\begin{adjustbox}{max width=.95\linewidth}
		\begin{tabularx}{\linewidth}{lX}
			\toprule
			\textbf{Test} & \textbf{Description} \\ 
			\midrule
			$\mathcal{T}_{L1}$ & EKF with real-world data, using navigation sensors and  visual markers.\\
			$\mathcal{T}_{L2}$ & $\mathcal{T}_{L1}$ plus visual odometry from plane registration (Sec.~\ref{sec: plane-based method}) and ORB-SLAM2 feature tracking (Sec.~\ref{sec: feature-based method}); selectively used based on IQA (Sec.~\ref{sec: adaptive_navigation_scheme}). \\
			$\mathcal{T}_{L3}$ & $\mathcal{T}_{L2}$ minus odometry from visual markers i.e., navigation sensors and image quality based VO inputs.\\
			\bottomrule
		\end{tabularx}
	\end{adjustbox}
\end{table}

\begin{table}[!b]
	\centering
	\footnotesize
	%\scriptsize
	%\small
	\captionsetup{justification=centering}
	\caption{Tests $\mathcal{T}_{Li}$ measure results for position/orientation error and trajectory autocorrelation}
	
	\begin{tabularx}{\linewidth}{@{}XYYY@{}}
		\toprule
		~ & $\mathcal{T}_{L1}$ & $\mathcal{T}_{L2}$ & $\mathcal{T}_{L3}$ \\ 
		\midrule
		$\bar{m}_{M,F}(\mathcal{T}_{Li}\langle \bar{\mathbf{p}} \rangle )\mathrm{[m]}$ & 0.65 $\pm$ 0.58 & 0.31 $\pm$ 0.11 & 0.85 $\pm$ 0.22 \\
		$\bar{m}_{M,F}(\mathcal{T}_{Li}\langle \bar{\mathbf{q}} \rangle )\mathrm{[deg]}$ & 14.65 $\pm$ 8.42 & 7.21 $\pm$ 2.10 & 11.89 $\pm$ 4.55 \\
		$m_{A}(\mathcal{T}_{Li})$ & 0.88 & 0.94 & 0.91 \\
		\bottomrule
	\end{tabularx}
	\label{table:localization_measures}
\end{table}

%In our previous work~\cite{Mueller2018_DexROVSIL}, we proved that the usage of visual landmarks substantially improves the localization filter accuracy compared to using only navigation sensors.
The use of visual landmarks has shown to substantially improve the localization filter accuracy~\cite{Mueller2018_DexROVSIL} compared to using only navigation sensors.
With data from DexROV sea trials, we first evaluate this method in $\mathcal{T}_{L1}$ and use it as reference.
Fig.~\ref{fig:results_localization_real_markers} shows that the majority of the largest errors occur when the robot is closer to the panel's corners because markers are observed from highly skewed perspectives.
Or when markers are not in view for a long period of time, e.g., in \emph{reference point 1} in Fig.~\ref{fig:results_localization_real_markers}.
Of course, there can be other sources of error like spurious DVL measurements that affect the overall accuracy.

In test $\mathcal{T}_{L2}$, we use our navigation scheme based on IQA.
%add all of our proposed visual odometry methods (planes registration and feature tracking) into the EKF. 
Table~\ref{table:localization_measures} and Fig.~\ref{fig:results_localization_real_markers_planes} show great reduction in the pose/orientation error $(\approx\%50)$ and an increase in the autocorrelation measure, i.e., smoother trajectories.
Moreover, errors at the panel's corners decrease, e.g. \emph{reference point 1}.
However, the largest errors still happen at these locations; after all, less features are observable and cylindrical corners (see Fig.~\ref{fig:rov_and_panel}) are imperfectly modeled by planes.

Finally in test $\mathcal{T}_{L3}$, we analyze the performance of our method without the use of visual landmarks.
The objective is to strive towards a more general localization filter that can function without fiducial landmarks. 
Table~\ref{table:localization_measures} shows that although the position and orientation error increase, they are not far from $\mathcal{T}_{L1}$ results. 
Furthermore, the error variance is significantly less; in Fig~\ref{fig:results_localization_real_plane} the circles representing the pose errors have a more uniform size.  
This is more suitable for control algorithms, i.e., waypoint navigation and manipulation, which need a certain response time to converge into desired states.
Highly variable measures may cause controllers to not converge. 
The same advantage can be said about the high autocorrelation values from $\mathcal{T}_{L2}$ and $\mathcal{T}_{L3}$.
In contrast, $\mathcal{T}_{L1}$ variances are more than $50\%$ of the mean error. 

\section{Conclusion}
\label{sec:conclusion}
Underwater operations are harsh due to the dynamic environment and the limited access to the system. 
However, the commercial demand to develop these technologies increases every year. 
One of the many challenges to tackle, and commonly the first in the work pipeline, is the achievement of robust, reliable and precise localization. 
For this reason, we investigate the use of visual odometry in underwater structured scenarios, especially a plane-based method adapted for underwater use with stereo processing and a standard feature based method. Furthermore, an image quality assessment is introduced that allows decision making to exclude computationally expensive visual processing, which is likely to lead to results with high uncertainty.  
%Although there is the possibility of always using man-made visual markers, the aim is to work towards a general framework that can be used in distinct environments and with the least human intervention as possible. 
The approach is validated in simulation and especially also in challenging field trials. 
%In this paper, we were able to validate and test with simulated and real-world data the proposed mechanisms; as well as to determine how different type of data impacts the accuracy, i.e., different image content characteristics and artifacts. 
%We believe that the achieved results have enough accuracy to enable other complex tasks such as underwater object manipulations. 
%Future work include the application of other surface primitives like cylinders or representation by superquadrics, as structured environments commonly contain simple geometrical shapes.  

\bibliographystyle{IEEEtran}
% argument is your BibTeX string definitions and bibliography database(s)
\bibliography{bibliography}

\end{document}